# Joint Maximum Purity Forest with Application to Image Super-Resolution

Hailiang Li[1], Kin-Man Lam[1], Dong Li[2]

[1]Department of Electronic and Information Engineering, The Hong Kong Polytechnic University
[2]School of Automation, Guangdong University of Technology

harley.li@connect.polyu.hk, enkmlam@polyu.edu.hk, dong.li@gdut.edu.cn

**Abstract** — In this paper, we propose a novel random-forest scheme, namely Joint Maximum Purity Forest (JMPF), for classification, clustering, and regression tasks. In the JMPF scheme, the original feature space is transformed into a compactly pre-clustered feature space, via a trained rotation matrix. The rotation matrix is obtained through an iterative quantization process, where the input data belonging to different classes are clustered to the respective vertices of the new feature space with maximum purity. In the new feature space, orthogonal hyperplanes, which are employed at the split-nodes of decision trees in random forests, can tackle the clustering problems effectively. We evaluated our proposed method on public benchmark datasets for regression and classification tasks, and experiments showed that JMPF remarkably outperforms other state-of-the-art random-forest-based approaches. Furthermore, we applied JMPF to image super-resolution, because the transformed, compact features are more discriminative to the clustering-regression scheme. Experiment results on several public benchmark datasets also showed that the JMPF-based image super-resolution scheme is consistently superior to recent state-of-the-art image super-resolution algorithms.

## I. Introduction

Recently, random forest [3, 14] has been employed as an efficient classification or regression tool on a large variety of computer-vision applications, such as object recognition [27], face alignment [15, 21, 46], data clustering [17], image super-resolution [8, 19], and so on. This method is attractive on computer-vision problems, not only for its simple implementation, but also for its many merits: (1) it can work efficiently on both the training and inference stages, (2) it is feasible for it to be sped up with parallel processing technology, (3) it has an inherent property to handle high-dimensional input features, and (4) it works with divide-and-conquer strategy, which has stable performance on classification and regression tasks as an ensemble machine-learning tool.

By studying the mechanism of a random forest, we can see that the random-forest approach has some critical properties, as do other powerful classifiers, such as SVM (support vector machine) [10, 48] and AdaBoost (short for "Adaptive Boosting") [13]. Both SVM and AdaBoost work as to approximate Bayes decision rule – known to be the optimal classifiers – via minimizing a margin-based global loss function. Each threshold in a decision tree of a random forest works as a hyperplane, and each single decision tree, similar to AdaBoost, attempts to minimize its global loss greedily and recursively on working through from the root-node down to leaf-nodes in the binary tree.

At each split-node in a decision tree, a hyperplane is learned to separate data into two groups. Although each decision tree attempts to achieve maximum purity for the two data groups clustered at each split-node independently during training a random forest, there is no guarantee that the original feature space can meet the expectation of global maximum purity for all the clustered groups. As the hyperplanes in a random forest have the orthogonal constraint, as shown in Fig. 1(b), which hinders us from achieving the optimal hyperplanes as SVM does (i.e., there is no orthogonal constraint in SVM) in some original feature space, as shown in Fig. 1(a). In this paper, we aim to solve this orthogonal-constraint limitation. With the fixed orthogonal hyperplanes, we propose to rotate the feature space, this is equivalent to rotating the hyperplanes, in such a way that global maximum purity on the clustered data can be achieved, as illustrated in Fig. 2. This strategy can achieve a joint maximum purity for all the split-nodes when training a random forest.

Image super-resolution can be performed based on clustering/classification, according to the recent emerging clustering-regression stream [2, 5, 8], and the JMPF scheme can achieve remarkable performance on both the classification and regression tasks. Therefore, JMPF is applied to single-image super-resolution in this paper. In our algorithm, principal component analysis (PCA) is applied to the features for dimensionality reduction. The projected feature space is then rotated to a compact, pre-clustered feature space via a learned rotation matrix. Finally, for all the split-nodes trained for a random forest, their thresholds are directly set to the inherent zero-center orthogonal hyperplanes in the rotated feature space to meet the maximum-purity criterion. Experiment results show that JMPF can achieve more accurate clustering/classification performance on random forests, and applying JMPF to image super-resolution can achieve superior quality, compared to state-of-the-art methods.

Having introduced the main idea of our proposed algorithm, the remainder of this paper is organized as follows. In Section II, we will describe our proposed scheme, the joint maximum purity forest scheme, and present in detail how to compute the rotation matrix via clustering data into the feature-space vertices. Section III will evaluate our proposed method and compare its performance with recent state-of-the-art random-forest-based approaches on regression and classification tasks. In Section IV, we will validate the performance of JMPF scheme on single-image super-resolution. Conclusions are given in Section V.

## II. JOINT MAXIMUM PURITY FOREST SCHEME

### II.1 Random Forest and Our Insights

A random forest is an ensemble of $T$ binary decision trees $\mathcal{T}^t(x): X \to \mathbb{R}^d$, where $t\ (= 1, 2, \dots, T)$ is the index of the trees, $X \in \mathbb{R}^m$ is the $m$-dimension feature space, and $\mathbb{R}^d = [0, 1]^d$ represents the space of class probability distributions over the label space $Y = \{1, \dots, d\}$. As shown in Fig. 1(b), the vertical dotted line forms a hyperplane, $X_1=0$, chosen in the first split-node for separating training samples, and the horizontal dotted line is the hyperplane, $X_2=0$, for the second split-node to cluster all the feature data assigned to this node. This results in separating the three data samples (Red, Green and Blue) into three leaf-nodes.

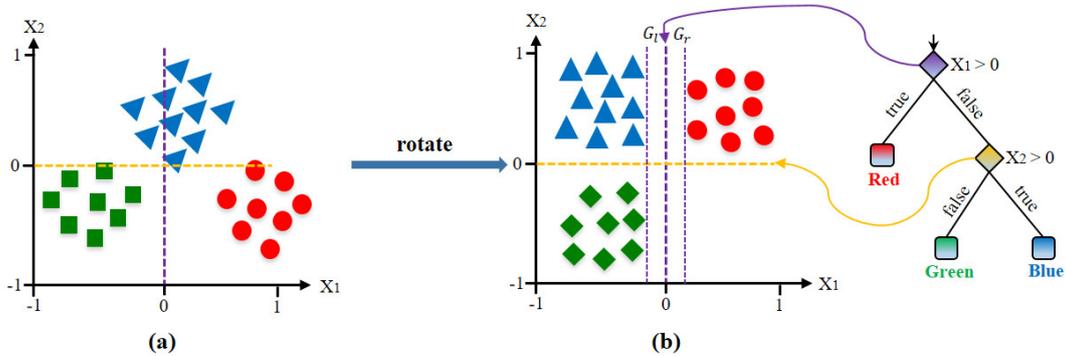

Fig. 1: (a) Three classes of samples in a feature space, which are hard to be clustered with orthogonal hyperplanes; and (b) the samples are rotated, and a decision tree of a random forest is used to cluster the data in the new, rotated feature space.

It can be seen from Fig. 1(b) that, for each split-node, the optimal hyperplane with more generalization capability is the one which can achieve maximum purity in clustering samples into two groups. For example, the vertical dotted line is the first optimal hyperplane because it clusters all the red training samples into the right node, while all the blue and green samples are clustered into the left node. Furthermore, the left margin $G_l$ and the right margin $G_r$ are equal. Although there is no guarantee that optimal hyperplanes can be determined for all the split-nodes in a random forest, approximated optimal hyperplanes can be obtained through a random bagging strategy.

The training of a whole random forest is to train all of its decision trees, by choosing the candidate features and thresholds for each of the split-nodes, where the feature dimensions and thresholds are determined using a random bagging strategy. In the prediction stage, each decision tree returns a class probability $p_t(y|x)$ for a given query sample $x \in \mathbb{R}^m$, and the final class label $y^*$ is then obtained via averaging, as follows:

$$y^* = \arg\max_y \frac{1}{T}\sum_{t=1}^{T} p_t(y|x). \qquad (1)$$

The splitting function for a split-node is denoted as $s(v;\Theta)$, where $v$ is a sample and $\Theta$ is typically parameterized by two values: (i) a feature dimension $\Theta^i \in \{1,\ldots,m\}$, and (ii) a threshold $\Theta^t \in \mathbb{R}$. The splitting function is defined as follows:

$$s(v;\Theta) = \begin{cases} 0, & \text{if } v(\Theta^i) < \Theta^t, \\ 1, & \text{otherwise,} \end{cases} \qquad (2)$$

where the outcome defines to which child node the sample $v$ is routed, and 0 and 1 are the two labels for the left and right child nodes, respectively. Each node chooses the best splitting function $\Theta^*$ out of a randomly sampled set $\{\Theta^t\}$ by optimizing the following function:

$$I = \frac{|L|}{|L|+|R|}H(L) + \frac{|R|}{|L|+|R|}H(R), \qquad (3)$$

where $L$ and $R$ are the sets of samples that are routed to the left and the right child nodes, and $|S|$ represents the number of samples in the set $S$. During the training of a random forest, the decision trees are provided with a random subset of the training data (i.e. bagging), and are trained independently of each other. Therefore, the decision trees are working as independent experts. Taking random-forest-based classification as an example, training a single decision tree involves recursively splitting each node, such that the training data in each newly created child node is clustered according to their corresponding class labels, so the purity at each node is increasing along a tree. Each tree is grown until a stopping criterion is reached (e.g. the number of samples in a node is less than a threshold or the tree depth reaches a maximum value) and the class probability distributions are estimated in the leaf-nodes. After fulfilling one of these criteria, a density model $p(y)$ in the leaf-node is estimated by all samples falling into this leaf-node for predicting the target value in the testing stage. A simple way to estimate the probability distribution $p(y)$ is averaging all the samples in the leaf-node, while there are also variant methods, such as fitting a Gaussian distribution or kernel density estimation, ridge regression [8, 21, 46], and so on.

$H(S)$ is the local score for a set of samples ($S$ is either $L$ or $R$), which normally is calculated using entropy as in Eqn. (4), but it can be replaced by variance [8, 21, 46] or the Gini index [14].

$$H(S) = -\sum_{k=1}^{K}[p(k|S) * \log(p(k|S))], \qquad (4)$$

where $K$ is the number of classes, and $p(k|S)$ is the probability for class $k$, given the set $S$. For the regression problem, the differential entropy:

$$H(q) = \int_y q(y|x) * \log(q(y|x))d_y \qquad (5)$$

over continuous outputs can be employed, where $q(y|x)$ denotes the conditional probability of a target variable given the input sample. Assuming $q(.,.)$ to be a Gaussian distribution and having only a finite set $S$ of samples, the differential entropy can be written in closed form as

$$H_{Gauss}(S) = \frac{K}{2}(1 - \log(2\pi)) + \frac{1}{2}\log(\det(\Sigma_S)), \qquad (6)$$

where $\det(\Sigma_S)$ is the determinant of the estimated covariance matrix of the target variables in $S$. For training each decision tree in a random forest, the goal on each split-node is to maximize the information gain (IG) by reducing the entropy after splitting. IG is defined as follows:

$$\text{IG} = \text{entropy(parent)} - [\text{average entropy(children)}]. \qquad (7)$$

Since each decision tree is a binary tree and each step is to split a current node (a parent set $S$) into two children nodes ($L$ and $R$ sets), IG can be described as follows:

$$\arg\max_{\mathcal{H}} IG = \arg\max_{L,R} H(S) - \frac{|L|}{|L|+|R|}H(L) - \frac{|R|}{|L|+|R|}H(R), \qquad (8)$$

where $\mathcal{H}$ is the optimal hyperplane of the split-node, and Eqn. (8) is the target function of each split-node when training each decision tree of a random forest. As we can see from Fig. 1(b), all the optimal hyperplanes from split-nodes are achieved independently and locally.

Since each optimal hyperplane is obtained from a subset of feature-dimension candidates with the randomly bagging strategy, there is no guarantee of obtaining a global optimum with respect to all the hyperplanes in all the split-nodes. An intuitive thinking, which was inspired by the data distribution in Fig. 1(b), is to achieve a global optimum by jointly considering all the hyperplanes of all the split-nodes, in the form as follows:

$$\max_{\mathcal{H}_k} IG_{global} = \arg\max_{\mathcal{H}_k} \prod_{k=1}^{\mathcal{K}} IG_k, \qquad (9)$$

where $\mathcal{K}$ is the total number of split-nodes that a training sample has routed through a decision tree. As there is no mathematical solution to the problem described in Eqn. (9), an alternative way (i.e., an approximate method) to numerically solving Eqn. (9) is to jointly maximize the purity of the clustered data groups at each of the split-nodes. This also means that all the data is clustered into the corners (feature-space vertices) of the feature space, as shown in Fig. 2.

**II.2 The Joint Maximum Purity Forest Scheme**

To calculate the threshold for each split-node in each decision tree when training a random forest, we are attempting to determine an orthogonal hyperplane for a three-category classification problem, as shown in Fig. 1. Since the hyperplanes for the split-nodes of a decision tree are required to be orthogonal to each other, seeking an optimal orthogonal hyperplane locally cannot guarantee obtaining maximum purity for the whole tree globally. As shown in Fig. 2, it is easy to determine the vertical hyperplane for maximum purity, but it is hard to obtain the horizontal hyperplane for maximum purity in the original feature space. To achieve an optimal classification performance for the whole decision tree, all the split-nodes should be considered globally or simultaneously.

As shown in Fig. 2, a number of split-nodes, which have their hyperplanes orthogonal to each other, are required to separate the samples into different nodes. However, if we can transform the samples (zero-centered feature data) to locate them at the respective corners of the feature space, i.e. $\{-1,1\}^m$ for $m$-dimensional features, the feature data can be easily and accurately separated by the orthogonal (either vertical or horizontal) hyperplanes, which contain the space center $\{0\}^m$, as illustrated in Fig. 1(b). The insight behind this is that the data is clustered into the feature-space vertices (the corners in a 2-D feature space means that the data points belong to $\{-1,1\}^2$ as the coordinate range is set to $[-1, 1]$).

To tackle the original feature data $X$, which is not ideally clustered in the vertices or corners of the feature space or close to them, as shown in Fig. 1(a), an intuitive idea is to rotate the feature space (this is equivalent to rotating the hyperplanes). This transformation clusters the feature data compactly into $m$ feature-space vertices $\{-1,1\}^m$ with a total of $2^m$ vertices. Therefore, a possible solution to the problem described in Eqn. (10) is to rotate the data features by a rotation matrix $\mathcal{R}^{m\times m}$, as shown in Fig. 2, through which the original feature space $X$ is transformed into a more compact clustered feature space, where all the feature data is clustered close to the feature-space vertices $B$. This solution can be mathematically defined as follows:

$$\min\|B - X\mathcal{R}\|_F^2, \text{s.t.} \ B \in \{-1,1\}^{n\times m}, \mathcal{R}^T\mathcal{R} = I \tag{10}$$

where $X \in \mathbb{R}^{n\times m}$ contains $n$ samples, each of which is a $m$-dimensional feature vector arranged in a row, and is zero-centered, i.e. all the feature vectors are demeaned by subtracting the mean vector from each feature vector.

This idea of clustering data into the feature-space vertices can also be found in locality-sensitive hashing (LSH) [1] and image representation [7]. In [1], a simple and efficient alternating minimization scheme was proposed to find a rotation matrix for zero-centered feature data, which minimizes the quantization errors by mapping the feature data to the vertices of a zero-centered binary hypercube. The method is termed as iterative quantization (ITQ), which can work on multi-class spectral clustering and orthogonal Procrustes problem. Yu et al. [54] proposed using a circulant matrix to speed up the computation, because the circulant structure enables the use of Fast Fourier Transformation (FFT). As the computation of the rotation matrix in the training and testing stage is ignorable, we choose a similar scheme to ITQ [1] to determine the rotation matrix $R$ (we throw away the final quantization matrix $B$ described in Eqn. (10), which is used for hashing in [1]), through which the original feature space $X$ can be transformed into a new compact clustered feature space: $\tilde{X} = X\mathcal{R}$, where the data is located at the respective vertices in the new feature space. After this transformation, a random forest with globally joint maximum purity of all the clustered data can be trained, through all the hyperplanes in the split-nodes of each decision tree. Based on this idea, our proposed scheme is called joint maximum purity forest (JMPF).

**II.3 Learning the Rotation Matrix via Clustering Data into Feature-Space Vertices**

Assuming that $x \in \mathbb{R}^m$ is one point in the $m$-dimensional feature space $X$(zero-centered data), the respective vertices in the zero-centered binary hypercube space can be denoted as $sgn(x) \in \{-1,1\}^m$, and there is a total of $2^m$ vertices in the $m$-dimensional feature space. It is easy to see from Fig. 2 that $sgn(x)$ is the vertex in the feature space, such that it is the closest to $x$ in terms of Euclidean distance. We denote a binary code matrix $B \in \{-1,1\}^{n\times m}$, whose rows $b = sgn(x) \in B$. For a matrix or a vector, $sgn(.)$ applies the sign operation to it element-wise.

Our objective is to minimize the error between the feature $X$ and the feature-space vertices $B$, i.e., $\min\|B - X\|^2$. As we can see in Fig. 2, when the feature space is rotated, the feature points will be more concentrated around their nearest vertices, which means that the quantization error will become smaller. Therefore, the minimization problem of $\min\|B - X\|^2$ is equivalent to minimizing the error of the zero-centered data with respect to the Frobenius norm, as in the following formulation:

$$Q(B, \mathcal{R}) = \|B - X\mathcal{R}\|_F^2, \text{s.t.} \ B \in \{-1,1\}^{n\times m}, \mathcal{R}^T\mathcal{R} = I. \tag{11}$$

Therefore, the task of this minimization problem is to determine an optimal rotation matrix $\mathcal{R}$ to satisfy Eqn. (11). Since there are two variables in Eqn. (11), the expectation–maximization (E-M) algorithm is applied to cluster data into the feature-space vertices, such that a local minimum of the binary code matrix $B$ and the rotation matrix $\mathcal{R}$ are computed simultaneously.

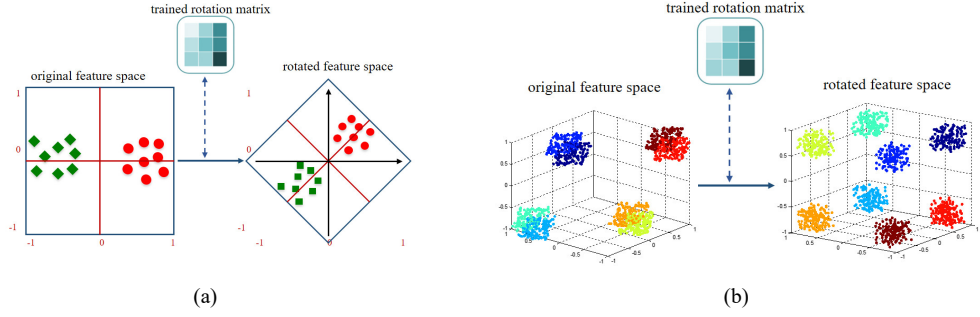

(a)          (b)

Fig. 2: Two toy examples of rotating a feature space into a more compact clustered feature space: (a) 2-dimensional features and (b) 3-dimensional features. The feature data is clustered into the vertices of a new feature space, by jointly maximizing the purity of all the clustered data.

The idea of rotating feature data to minimize the error between the transformed data and the feature-space vertices $B$ can also be found in [7], which showed that the rotation matrix $\mathcal{R}$ can be initialized randomly, and then iterated to converge to the required rotation matrix. Two iteration steps will be performed: in every iteration, each feature vector in the feature space is firstly quantized to the nearest vertex of the binary hypercube, i.e. to a vertex in $B$, and then the rotation matrix $\mathcal{R}$ is updated to minimize the quantization error by fixing $B$. These two alternating steps are described in detail below:

(1) *Fix $\mathcal{R}$ and update $B$*:

$$\begin{aligned} Q(B,\mathcal{R}) &= \|B - X\mathcal{R}\|_F^2 \\ &= \|B\|_F^2 + \|X\|_F^2 - 2tr(B\mathcal{R}^T X^T) \\ &= n \times m + \|X\|_F^2 - 2tr(B\mathcal{R}^T X^T) \end{aligned} \quad (12)$$

Because the zero-centered data matrix $X$ is fixed, minimizing Eqn. (12) is equivalent to maximizing the following term:

$$tr(B\mathcal{R}^T X^T) = \sum_{i=1}^{n}\sum_{j=1}^{m} B_{ij} \tilde{X}_{ij} \quad (13)$$

where $\tilde{X}_{ij}$ is an element of $\tilde{X} = X\mathcal{R}$. To maximize Eqn. (13) with respect to $B$, $B_{ij} = 1$ whenever $\tilde{X}_{ij} \geq 0$ and $B_{ij} = -1$ otherwise, i.e. $B = sgn(X\mathcal{R}) \in \{-1,1\}^m$.

(2) *Fix $B$ and update $\mathcal{R}$*:

The problem of fixing $B$ to obtain a rotation matrix based on the objective function Eqn. (11) is relative to the classic orthogonal Procrustes problem [6, 34, 55], in which a rotation matrix is determined to align one point set with another.

In our algorithm, these two point sets are the zero-centered data set $X$ and the quantized matrix $B$. Therefore, a closed-form solution for $\mathcal{R}$ is available, by applying SVD on the $m \times m$ matrix $B^T X$ to obtain $S\Omega \hat{S}^T$ ($\Omega$ is a diagonal matrix), then set $\mathcal{R} = \hat{S}S^T$ to update $\mathcal{R}$.

**II.4 Proof of the Orthogonal Procrustes Problem:**

For completeness, we prove the orthogonal Procrustes problem, for which the solution can be found in [6, 34, 55]:

Problem definition: $\quad\quad \min_{\mathcal{R}} \|B - X\mathcal{R}\|_F^2 \quad$ subject to: $\mathcal{R}^T\mathcal{R} = I$ . $\quad\quad (14)$

Proof: $\quad\quad \|B - X\mathcal{R}\|_F^2 \quad\quad (15)$
$$= tr(B - X\mathcal{R})(B^T - \mathcal{R}^T X^T)$$
$$= tr(BB^T) - 2tr(BX^T\mathcal{R}^T) + tr(\mathcal{R}XX^T\mathcal{R}^T)$$

Thus, $\min_{\mathcal{R}}\|B - \mathcal{R}X\|_F^2$ equals to maximizing:

$$
\begin{aligned}
tr(BX^T\mathcal{R}^T) \quad & ([U,\Omega,V] = svd(BX^T)) \\
& = tr(U\Omega XV^T\mathcal{R}^T) \\
& = tr(\Omega V^T\mathcal{R}^T U) \quad (Z = V^T\mathcal{R}^T U) \\
& = tr(\Omega Z) \\
& = tr \sum_i Z_{i,i}\Omega_{i,i} \\
& \leq \sum_i \Omega_{i,i}
\end{aligned}
\tag{16}
$$

The last inequality holds because Z is also an orthonormal matrix, and $\sum_j Z_{i,j}^2 = 1, Z_{i,i} \leq 1$. The objective function can be maximized if $Z = I$, i.e.

$$\mathcal{R} = UV^T \qquad \blacksquare$$

### III. JOINT MAXIMUM PURITY FOREST FOR REGRESSION AND CLASSIFICATION

**III.1 The Workflow of Joint Maximum Purity Forest**

Random forest is a machine-learning method using an ensemble of randomized decision trees for classification. Each tree in a random forest consists of split-nodes and leaf-nodes, which can be trained recursively. A random forest is constructed recursively, where each node attempts to find a splitting function or a hyperplane to separate its samples into two leaf-nodes, such that the information gain is optimized. A tree stops growing if the maximum depth is reached or if a node has achieved maximum purity, i.e. it contains only samples from one class. Then, each leaf-node collects the statistics of the samples falling in it. In the evaluation phase, the probability of a query sample *x* belonging to class *k* is given by averaging all the trees, or by other methods.

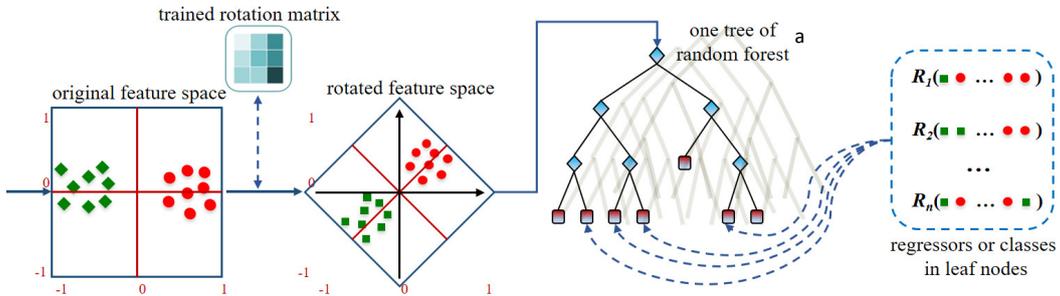

Fig. 3: An overview of the workflow of the JMPF-based random forest.

Most random-forest-based models [8, 21, 23, 24] share a similar workflow, as shown in Fig. 3, in which the main task on training a tree in a random forest is to decide thresholds in the split-nodes and learn the regressors or classes in the leaf-nodes. Rigid regression or linear regression is often employed in the leaf-nodes for the prediction task, because rigid regression has a closed-form solution, while linear regression is an efficient optimization tool, and the *LibLinear* package [53] can be used to fine-tune its configurations.

Compared to conventional random forests, our JMPF scheme has one more step, as shown in the left of Fig. 3, the rotation matrix. The JMPF scheme transforms the original feature space by rotating it into a more compact, pre-clustered feature space, using a trained rotation matrix learned through clustering feature vectors iteratively into the vertices of a new feature space. The whole workflow of our proposed algorithm, the JMPF scheme, is outlined in Fig. 3. The source code of our algorithm is available to download at: https://github.com/HarleyHK/JMPF.

**III.2 The inherent zero-center hyperplanes as thresholds for split-nodes**

In training a random forest, the two main operations for training (splitting) each split-node are to choose splitting feature(s), and to determine the threshold, using a random bagging strategy, which can avoid over-fitting in training classifiers. In the rotated compact pre-clustered feature space, the inherent zero-center hyperplanes are inherently the optimal thresholds (to meet the max-purity criterion on two clustered data groups) after training the rotation matrix. Therefore, these inherent zero-center hyperplanes can directly be set as the thresholds to achieve optimal classification performance on training a random forest. Compared to conventional random forests, our proposed JMPF only needs to choose which feature(s) to split data at split-nodes. This can speed up the training process for a random forest. Experimental results in the next subsection will validate this performance.

**III.3: Experimental results on JMPF regression and classification**

To evaluate the performances of the proposed JMPF, we test it with 15 standard machine-learning tasks, 7 for classification and 8 for regression. The datasets used in the experiments are summarized in Table-1. We use standard performance evaluation metrics: error rate for classification and root mean squared error (RMSE) for regression, unless otherwise specified.

| Dataset | #Train | #Test | #Feature | #Classes or TargetDim |
|---|---|---|---|---|
| (c)char74k | 66707 | 7400 | 64 | 62 |
| (c)gas sensor | 11128 | 2782 | 128 | 6 |
| (c)isolet | 6238 | 1558 | 617 | 26 |
| (c)letterorig | 16000 | 4000 | 16 | 26 |
| (c)pendigits | 7494 | 3498 | 16 | 10 |
| (c)sensorless | 46800 | 11700 | 48 | 11 |
| (c)usps | 7291 | 2007 | 256 | 10 |
| (r)delta ailerons | 7129*3/4 | 7129/4 | 5 | 1 |
| (r)delta elevators | 5720 | 3807 | 6 | 1 |
| (r)elevators | 8752 | 7847 | 18 | 1 |
| (r)kin8nm | 8192*3/4 | 8192/4 | 8 | 1 |
| (r)price | 159*3/4 | 159/4 | 15 | 1 |
| (r)pyrim | 74*3/4 | 74/4 | 27 | 1 |
| (r)stock | 950*3/4 | 950/4 | 10 | 1 |
| (r)WiscoinBreastCancer | 194*3/4 | 194/4 | 32 | 1 |

Table-1: The properties of the standard machine-learning datasets used for classification and regression. The top 7 are used for classification (c) and the bottom 8 for regression (r). (3/4 means 75% training and 25% testing)

We firstly evaluate the proposed approach on two real applications, one for classification (Table-2) and one for regression (Table-3). Our proposed JMPF is compared with the original random forest before refinement (denoted as RF), and two state-of-the-art variants: alternating decision forests (ADF) [23] and alternating regression forests (ARF) [24], for classification and regression, respectively. Furthermore, we compare with JMPF+ADF/ARF, for demonstrating that our algorithm can be combined with other methods. We follow the experiment settings in [23, 24]. We set the maximum tree depth $D$ at 15, and the minimum sample number in a splitting node is set at 5. The experiments were repeated five times, and the average error and standard deviation were measured. The results are presented in Table-2 and Table-3, for the classification and regression tasks, respectively. In terms of accuracy, our proposed JMPF significantly outperforms the standard random forest on all classification and regression tasks. Compared to RF, JMPF achieves an average of 23.57% improvement on the classification tasks, and an average of 23.13% improvement on the regression tasks. Our method also consistently outperforms the state-of-the-art variants: ADF/ARF. Moreover, the performance of our JMPF algorithm can be further improved by integrating with ADF and ARF, denoted as JMPF + ADF/ARF. As shown in Table-2 and Table-3, JMPF+ADF achieves an average 27.86% improvement on the classification tasks, while JMPF+ARF achieves an average 26.88% improvement on the regression tasks. These results on diverse tasks clearly demonstrate the effectiveness of our proposed approach.

| dataset | #$\mathcal{H}$ | RF | ADF | JMPF | JMPF+ADF | $\lambda$ |
|---|---|---|---|---|---|---|
| char74k | 1 | 2.261±0.021 | 2.173±0.014 | **2.147**±0.021 **(05%)** | 2.114±0.016 **(07%)** | |
| | 3 | 2.449±0.029 | 2.236±0.015 | **2.206**±0.027 **(10%)** | 2.143±0.024 **(12%)** | $10^{-1}$ |
| | 5 | 2.452±0.016 | 2.232±0.021 | **2.209**±0.019 **(10%)** | 2.138±0.017 **(13%)** | |
| gas sensor | 1 | 5.656±0.534 | 5.238±0.539 | **4.211**±0.252 **(26%)** | 3.958±0.508 **(30%)** | |
| | 3 | 6.264±0.042 | 5.952±0.323 | **4.622**±0.299 **(26%)** | 4.416±0.370 **(30%)** | $10^{-3}$ |
| | 5 | 6.470±0.332 | 5.751±0.792 | **4.775**±0.459 **(26%)** | 4.159±0.324 **(36%)** | |
| isolet | 1 | 6.932±0.281 | 6.208±0.338 | **6.153**±0.381 **(11%)** | 5.868±0.239 **(15%)** | |
| | 3 | 6.501±0.199 | 6.308±0.330 | **6.272**±0.332 **(04%)** | 5.932±0.177 **(09%)** | $10^{-2}$ |
| | 5 | 7.005±0.362 | 6.528±0.261 | **6.381**±0.254 **(09%)** | 5.969±0.205 **(15%)** | |
| letterorig | 1 | 6.371±0.099 | 4.418±0.082 | **4.114**±0.087 **(35%)** | 3.535±0.111 **(45%)** | |
| | 3 | 6.889±0.199 | 5.196±0.127 | **4.864**±0.267 **(29%)** | 4.146±0.192 **(40%)** | $10^{-2}$ |
| | 5 | 6.739±0.263 | 5.082±0.097 | **4.625**±0.257 **(31%)** | 4.032±0.131 **(40%)** | |
| pendigits | 1 | 3.528±0.124 | 3.234±0.106 | **2.912**±0.069 **(17%)** | 2.850±0.136 **(19%)** | |
| | 3 | 3.418±0.171 | 3.377±0.164 | **2.969**±0.120 **(13%)** | 2.915±0.100 **(15%)** | $10^{-2}$ |
| | 5 | 3.499±0.184 | 3.283±0.184 | **3.054**±0.081 **(13%)** | 3.002±0.086 **(14%)** | |
| sensorless | 1 | 1.824±0.018 | 0.972±0.028 | **0.324**±0.005 **(82%)** | 0.253±0.009 **(86%)** | |
| | 3 | 1.026±0.158 | 0.391±0.007 | **0.293**±0.004 **(71%)** | 0.281±0.003 **(73%)** | $10^{-1}$ |
| | 5 | 0.903±0.150 | 0.512±0.223 | **0.268**±0.054 **(70%)** | 0.244±0.029 **(73%)** | |
| usps | 1 | 6.128±0.181 | 6.149±0.208 | **6.085**±0.216 **(01%)** | 5.964±0.206 **(03%)** | |
| | 3 | 6.527±0.203 | 6.520±0.188 | **6.285**±0.101 **(04%)** | 6.206±0.245 **(05%)** | $10^{-2}$ |
| | 5 | 6.548±0.225 | 6.441±0.195 | **6.391**±0.063 **(02%)** | 6.213±0.112 **(05%)** | |

Table-2: Comparison of classification performances on seven datasets, which can be found at UCI machine-learning repository: https://archive.ics.uci.edu/ml/datasets.html. RF: standard random forest, ADF: alternating decision forests [23], JMPF: proposed algorithm, JMPF+ADF: our proposed algorithm embedded into ADF. #$\mathcal{H}$ is the number of randomly chosen hyperplane(s) on training a random forest. $\lambda$ is the error scale. The percentages in brackets for JMPF and JMPF+ADF are the reduction rates in RMSE (root mean squared error) compared with the RF algorithm.

| dataset | RF | ARF | JMPF | JMPF+ARF | $\lambda$ |
|---|---|---|---|---|---|
| delta ailerons | 2.970±0.001 | 2.967±0.006 | **1.952**±0.003 **(34%)** | 1.946±0.002 **(34%)** | $10^{-4}$ |
| delta elevators | 2.360±0.002 | 2.338±0.008 | **1.635**±0.001 **(30%)** | 1.610±0.006 **(32%)** | $10^{-3}$ |
| elevators | 0.638±0.001 | 0.635±0.001 | **0.619**±0.001 **(03%)** | 0.606±0.001 **(05%)** | $10^{-2}$ |
| kin8nm | 2.622±0.002 | 2.545±0.003 | **1.962**±0.003 **(25%)** | 1.667±0.005 **(36%)** | $10^{-1}$ |
| price | 7.281±0.755 | 6.663±0.794 | **5.460**±0.627 **(25%)** | 5.234±0.666 **(28%)** | $10^{1}$ |
| pyrim | 1.440±0.008 | 1.042±0.347 | **1.031**±0.017 **(28%)** | 0.631±0.018 **(56%)** | $10^{-1}$ |
| stock | 2.878±0.022 | 2.823±0.038 | **2.744**±0.019 **(05%)** | 2.678±0.021 **(07%)** | $10^{0}$ |
| Wiscoin breast cancer | 3.669±0.041 | 3.130±0.044 | **3.081**±0.008 **(16%)** | 3.036±0.023 **(17%)** | $10^{1}$ |

Table-3: Comparison of regression performances on eight datasets, which can be found at http://www.dcc.fc.up.pt/~ltorgo/Regression/DataSets.html. RF: standard random forest, ARF: alternating regression forests [24], JMPF: proposed algorithm, JMPF+ARF: our proposed algorithm embedded into ARF. $\lambda$ is the error scale. The number of randomly chosen hyperplanes #$\mathcal{H}$ is 3. The percentages in brackets for JMPF and JMPF+ARF are the reduction rates in RMSE compared with the RF algorithm.

### III.4: Discussions on Experimental Results

The computational complexity of JMPF is similar to that of the standard random forest. As illustrated in the workflow of JMPF in Fig. 3, only one additional step, which computes the rotation matrix, is required, when compared to the standard random forest. For a small dataset (e.g., feature dimension size less than 500 and data size less than 10,000), the computation required to compute the rotation matrix for clustering data into the feature-space vertices is acceptable in the training stage (about 10 seconds per level, using MatLab) and negligible in the testing stage. When the dimension size becomes larger, PCA dimensionality reduction can be employed. If the size of the dataset increases, such that using PCA still involves heavy computation, bagging can be used to achieve comparable accuracy and the whole extra computation will be insignificant.

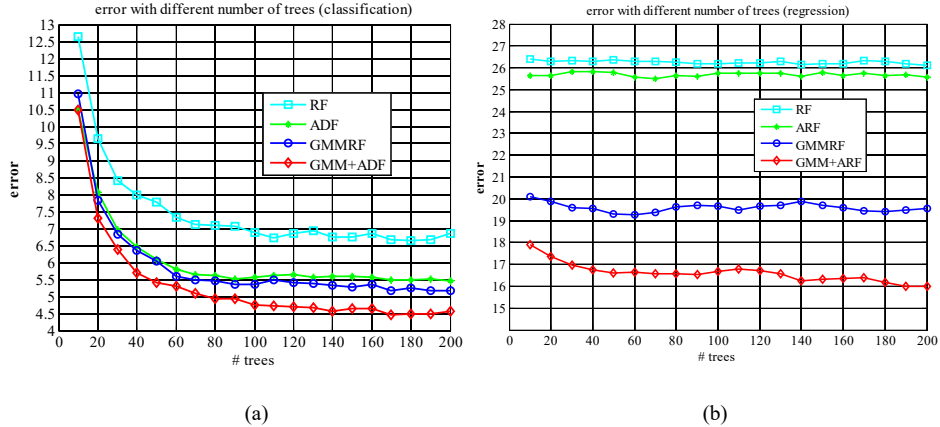

Fig. 4: Performance with different numbers of trees for (a) classification and (b) regression (dataset for classification is *letterorig* and dataset for regression is *kin8nm*, error scale: $10^{-2}$, the number hyperplane(s) #$\mathcal{H}$ on training the random forest is 3).

To study the stability of JMPF, we choose the *letterorig* dataset for classification and the *kin8nm* dataset for regression, and the respective results are shown in Fig. 4(a) and Fig. 4(b), respectively. In the experiments, the number of trees, i.e., the number of weak classifiers in the random forest, varies from 10 to 200, and we have three observations. Firstly, as shown in Fig. 4, when the number of trees increases, the performance of all the algorithms improves. For classification, as shown in Fig. 4(a), when the number of trees is larger than 100, the errors are converged to become steady. On the contrary, for the regression task as shown in Fig. 4(b), the errors are almost stable, ranged from 10 to 200. Secondly, the results show that JMPF consistently outperforms ADF and RF, irrespective of the number of trees used. Finally, Fig. 4 clearly shows that JMPF can integrate with ADF or ARF to further improve its performance.

## IV. IMAGE SUPER-RESOLUTION BASED ON JOINT MAXIMUM PURITY FOREST

### IV.1 Overview of Image Super-resolution and Related Works

Image super-resolution (SR), which recovers a high-resolution (HR) image from one single image or a number of low-resolution (LR) images, has been a hot research topic in the field of image processing for decades. SR is a well-known ill-posed problem, which needs artistic skills from mathematics and machine learning. Prior methods on SR are mainly based on edge preserving, such as New Edge-directed Interpolation (NEDI) [49], Soft-decision Adaptive Interpolation (SAI) [50], Directional Filtering and Data-Fusion (DFDF) [51], Modified Edge-Directed Interpolation (MEDI) [52], etc.

The neighbor-embedding (NE) methods [29, 30] set the milestone on the patch-learning-based super-resolution approach. In this approach, each LR patch is approximated as a linear regression of its nearest LR neighbors in a collected dataset, while its HR counterpart can be reconstructed with the same coefficients of corresponding HR neighbors, based on the non-linear manifold structure. Although the NE method is simple and practical, it requires a huge dataset (millions of patches) to achieve good reconstruction quality and it is computationally intensive, because *k*-NN is used in searching neighboring patches in the huge dataset. Instead of using the patches extracted directly from natural images, Yang *et al*. [28] employed sparse coding [12, 28] to represent patch images, of large size, efficiently, which opens the era for sparse coding in the image inverse problems.

The sparse-coding super-resolution (ScSR) approach is a framework that the HR counterpart of an LR patch can be reconstructed aided by two learned dictionaries, with the sparse constraint on the coefficients via the following formulations:

$$y \approx D_l\alpha, \quad x \approx D_h\alpha, \quad \alpha \in \mathbb{R}^k \quad \text{with} \quad \|\alpha\|_0 \ll k. \tag{17}$$

The compact LR and HR dictionaries can be jointly learned with a sparsity constraint, using the following sparse representation:

$$D_h, D_l = \underset{D_h, D_l}{\operatorname{argmin}} \|x - D_h\alpha\|_2^2 + \|y - D_l\alpha\|_2^2 + \lambda\|\alpha\|_0, \tag{18}$$

where $y$ and $x$ are the LR patch and the corresponding HR patch, respectively; and $D_l$ and $D_h$ are the LR and HR dictionaries learned from the LR and the corresponding HR patch samples, respectively. The value of $k$ in $\|\alpha\|_k$ is the sparsity factor of the coefficients $\alpha$. $\|\alpha\|_0$ is $l^0$-norm, which means the non-zero count of the coefficients in $\alpha$. For each LR patch $y$ of an input LR image $Y$, the problem of finding the sparse coefficients $\alpha$ can be formulated as follows:

$$\min\|\alpha\|_0 \quad \text{s.t.} \quad \|D_l\alpha - y\|_2^2 \leq \varepsilon \tag{19}$$

or

$$\min\|\alpha\|_0 \quad \text{s.t.} \quad \|FD_l\alpha - Fy\|_2^2 \leq \varepsilon, \tag{20}$$

where $F$ is a linear or non-linear feature-extraction operator on the LR patches, which makes the LR patches more discriminative from each other. Typically, $F$ can be chosen as a high-pass filter, and a simple high-pass filter can be obtained by subtracting the input from the output of a low-pass filter, as in an early work [44]. In [2, 4, 5, 28], first and second-order gradient operators are employed on up-sampled versions of low-resolution images, then four patches are extracted from these gradient maps at each location, and concatenate them to become feature vectors. The four 1-D filters used to extract the derivatives are:

$$\left. \begin{array}{l} F_1 = [-1, 0, 1], \ F_2 = F_1^T \\ F_3 = [1, 0, -2, 0, 1], \ F_4 = F_3^T \end{array} \right\} \tag{21}$$

The ideal regularization term for the sparse constraint on the coefficients α is the $l^0$-norm (non-convex), but, based on greedy matching, it leads to an NP-hard problem. Alternatively, Yang et al. [28] relaxed it to $l^1$-norm, as shown in the following formulation:

$$\min\|\alpha\|_1 \quad \text{s.t.} \quad \|FD_l\alpha - Fy\|_2^2 \leq \varepsilon. \tag{22}$$

The Lagrange multiplier provides an equivalent formulation as follows:

$$\min_\alpha \|FD_l\alpha - Fy\|_2^2 + \lambda\|\alpha\|_1, \tag{23}$$

where the parameter $\lambda$ balances the sparsity of the solution and the fidelity of the approximation to $y$. However, the effectiveness of sparsity was challenged in [5, 9], as to whether real sparsity can help image classification and restoration, or locality property can achieve the same effect. Timofte et al. [2] proposed an anchored neighborhood regression (ANR) framework, which relaxes the sparse decomposition optimization ($l^1$-norm) of [4, 28] to a ridge regression ($l^2$-norm) problem.

An important step in the ANR model is the relaxation of the $l^1$-norm in Eqn. (23) to the $l^2$-norm least-squares minimization constraint, as follows:

$$\min_\alpha \|FD_l\alpha - Fy\|_2^2 + \lambda\|\alpha\|_2, \tag{24}$$

where $D_l$ and $D_h$ are the LR and HR patch-based dictionaries, respectively. This $l^2$-norm constraint problem can be solved with a closed-form solution from the ridge regression [16] theory. Based on the Tikhonov regularization/ridge-regression theory, the closed-form solution of the coefficients is given:

$$\alpha = (D_l^T D_l + \lambda I)^{-1} D_l^T F y. \tag{25}$$

We assume that the HR patches share the same coefficient $\alpha$ from their counterpart LR patches, i.e., $x = D_h \alpha$. From Eqn. (25), we have:

$$x = D_h(D_l^T D_l + \lambda I)^{-1} D_l^T F y. \tag{26}$$

Therefore, the HR patches can be reconstructed by: $x = P_G F y$, where $P_G$ can be considered a projection matrix, which can be calculated offline, as follows:

$$P_G = D_h(D_l^T D_l + \lambda I)^{-1} D_l^T. \tag{27}$$

Ridge regression allows the coefficients $\alpha$ to be calculated by multiplying the constant projection matrix $P_G$ with the new extracted feature $Fy$, as described in Eqn. (26) and Eqn. (27). More importantly, the projection matrix $P_G$ can be pre-computed, and this offline learning enables significant speed-up at the prediction stage.

Timofte *et al*. [5] further extended the ANR approach to the A+ approach, which learns regressors from all the training samples, rather than from a small quantity of neighbors of the anchor atoms as ANR does. Later, there are numerous variants and extended approaches, based on ANR and A+ [9, 18, 22, 33, 35, 36, 45, 47]. By investigating the ANR model, Li et al. [9] found that the weights of the supporting atoms can be of different values to represent their similarities to the anchor atom. Based on this idea, the normal collaborative representation (CR) model in ANR is generalized to a weighted model, named as weighted collaborative representation (WCR) model, as follows:

$$\min_{\alpha} \|FD_l \alpha - Fy\|_2^2 + \|\lambda_{WCR}\alpha\|_2, \tag{28}$$

where $\lambda_{WCR}$ is a diagonal matrix. The weights on the diagonal atoms are proportional to their similarities to the anchor atom. Similarly, the new closed-form solution for the coefficients can be calculated offline, as follows:

$$\alpha^* = (D_l^T D_l + \lambda_{WCR})^{-1} D_l^T F y, \tag{29}$$

and the new projection matrix is given as follows:

$$P_G^* = D_h(D_l^T D_l + \lambda_{WCR})^{-1} D_l^T. \tag{30}$$

The WCR model can further improve the ANR or A+ model in terms of image quality, but it is still a time-consuming problem to find the most similar anchor atoms in a dictionary, and this always hinders its applications where fast speed is greatly required.

Schulter *et al*. [8] adopted the random forest as a classifier, and the regressors are learned from the patches in the leaf-nodes. With the same number of regressors, these random-forest-based methods [8, 41, 42, 43] can perform on a par with the A+ method in terms of accuracy. However, they achieve an increase in speed, because the sublinear search property of random forest can remarkably reduce the regressors' search complexity.

Recently, deep learning has become a hot research topic, which has been successfully applied to image super-resolution [37, 38, 39, 40] and achieved promising performance, particularly in terms of image quality. In [37, 38], a convolutional neural-network-based image super-resolution (SRCNN) was proposed, in which an end-to-end mapping between LR and HR images is learned through a deep convolutional neural network (CNN). [39] presented a super-resolution approach with very deep networks with extremely high learning rates, and the deep network convergence rate is sped up by residual learning. Meanwhile, [40] presented a generative adversarial network (GAN)-based deep residual network model for image super-resolution (SRGAN), in which content loss and adversarial loss are combined as an image perceptual loss function. The proposed deep residual network in [40] can super-resolve photo-realistic textures from 4-times down-sampled images, and an extensive mean-opinion-score (MOS) criterion is proposed to test the perceptual quality gained by using the SRGAN approach. Although deep-learning-based approaches can achieve superior performance compared to other SR methods, their heavy computation is always a big obstacle to their extensive applications with real-time requirements, where the graphics processing unit (GPU) may not be available, such as smart mobile phones.

**IV.2 JMPF-based Image Super-Resolution**

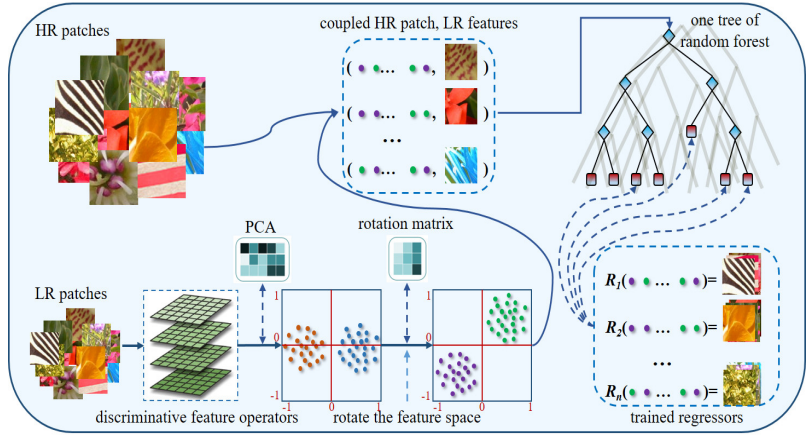

Fig. 5: An overview of the training process of the JMPF-based method for image super-resolution.

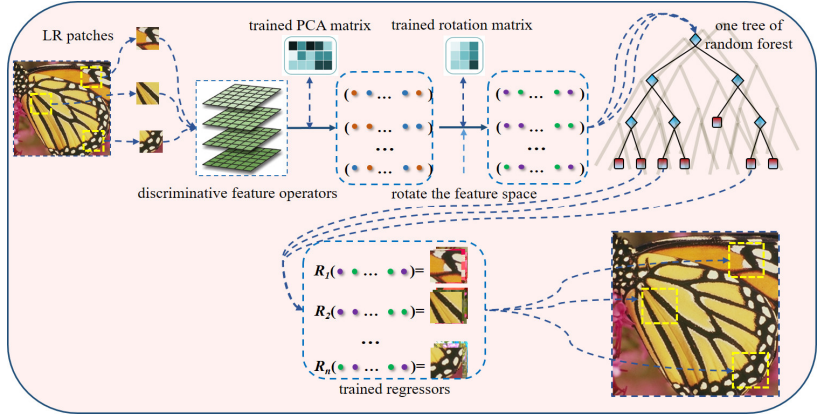

Fig. 6: An overview of the testing process of the JMPF-based method for image super-resolution

The recent emerging stream [5, 31] on single-image SR is to formulate the problem as a clustering-regression problem, which can be solved with machine-learning tools. These approaches are learning-based methods, which attempt to reconstruct an HR image from patches with the help of an external database. These methods first decompose an image into patches, then classify them into clusters. Regressors are then trained for each of the clusters, which generate mappings from an input LR patch's feature to its corresponding HR patch (see Fig. 5). In the testing stage, an LR query image follows the same procedures to cut into patches and to extract features, which are then assigned to their corresponding clusters using the *k*-NN algorithm [8, 19] or random forest [2, 5, 7]. The respective HR patches are constructed through regressors learned for the clusters (see Fig. 6). This kind of clustering-regression algorithms, based on random forest [2, 5, 7], has achieved state-of-the-art performance in single image super-resolution, both in terms of accuracy and efficiency, because of the use of ensemble learning and sublinear search. As JMPF achieves promising results on both classification and regression tasks, it can be employed for image super-resolution for better performances.

An overview of the training and testing processes of the proposed JMPF-based image SR method is illustrated in Fig. 5 and Fig. 6, respectively. In our method, the first and second-order gradients are extracted as features from each patch, followed by PCA for dimensionality reduction. These features are then rotated into a more compact, pre-clustered feature space. Finally, all the thresholds are directly set to the inherent zero-center hyperplanes when training the random forest, and similar to other algorithms, the regressors at the leaf-nodes are computed using the rigid regression algorithms. This approach is named as JMPF-based image super-resolution method.

### IV.3 The Working Processes of JMPF-based Image Super-resolution

JMPF has been shown to achieve a better performance for clustering and classification than other random forest methods. Since image super-resolution can be considered as a clustering/classification problem, using JMPF is likely to result in better performance. This is mainly due to the features transformed to the vertices in the new feature space, so the features become more discriminative. The image super-resolution training and testing processes of our proposed JMPF-based method are described in Algorithm 1 and Algorithm 2, respectively.

---

**Algorithm 1: JMPF-based Image Super-Resolution Training Process:**

**Input:** $\{x_i^l, x_i^h\}_{i=1}^N$: training LR-HR patch pairs, $N$ is the number of training samples.

**Output:** the random forest and ridge regression projection matrices: $\wp = (P_1, ..., P_T)$, in leaf-nodes, where $T$ is the number of regressors; the PCA projection matrix $\mathcal{M}$ and the rotation matrix $\mathcal{R}$.

**1:** Discriminative features calculated from patch images based on first and second-order (horizontal and vertical) gradients; ⇒ {Eqn. (21)}

**2:** Apply PCA on features to compute the PCA projection matrix $\mathcal{M}$;

**3:** Train a JMPF-based random forest by clustering PCA projected feature data into feature-space vertices, which can rotate the feature space into a compact pre-clustered feature space, at the same time obtain the rotation matrix $\mathcal{R}$; ⇒ {Eqn. (11)}

**4:** Train ridge regression projection matrices: $\wp = (P_1, ..., P_T)$, from LR-HR patch pairs in all the leaf-nodes. ⇒ {Eqn. (27)}

---

**Algorithm 2: JMPF-based Image Super-Resolution Testing Stage:**

**Input:** testing LR image $I^l$, the trained JMPF-based random forest and ridge regression projection matrices: $\wp = (P_1, ..., P_T)$ in leaf-nodes; the trained PCA projection matrix $\mathcal{M}$ and the trained rotation matrix $\mathcal{R}$.

**Output:** super-resolved image $I^h$.

**1:** Extract discriminative features for all the patches of image $I^l$; ⇒ {Eqn. (21)}

**2:** Do feature dimension reduction via the PCA projection matrix $\mathcal{M}$;

**3:** Rotate feature space into a compact pre-clustered feature space via the rotation matrix $\mathcal{R}$;

**4:** For LR patches from image $I^l$, based on their features, searching their corresponding regressors from leaf-nodes in the trained random-forest;

**5:** Produce $I^h$ through all the image patches from image $I^l$ by ridge regression with the trained projection matrices: $\wp = (P_1, ..., P_T)$. ⇒ {Eqn. (26)}

## IV.4 Experimental Results on JMPF-based Image Super-Resolution

In this section, we evaluate our image SR algorithm on some standard super-resolution datasets, including Set 5, Set14, and B100 [20], and compare it with a number of classical or state-of-the-art methods. These include bicubic interpolation, sparse representation SR (Zeyde) [4], anchored neighborhood regression (ANR) [2], A+ [5], standard random forest (RF) [8], and alternating regression forests (ARF) [8]. We set the same parameters for all the random-forest-based algorithms: the number of trees in the random forest is 10, and the maximum depth of each tree is 15.

| Dataset | *scale* | bicubic | Zeyde[4] | ANR[2] | A+[5] | RF[8] | ARF[8] | JMPF⁻ | JMPF | **JMPF⁺** |
|---|---|---|---|---|---|---|---|---|---|---|
| **Set5** | ×2 | 33.66 | 35.78 | 35.83 | 36.55 | 36.52 | **36.65** | 36.53 | 36.59 | *36.70* |
|  | ×3 | 30.39 | 31.92 | 31.93 | **32.59** | 32.44 | 32.53 | 32.51 | **32.59** | *32.67* |
|  | ×4 | 28.42 | 29.74 | 29.74 | **30.28** | 30.10 | 30.17 | 30.14 | 30.17 | *30.24* |
| **Set14** | ×2 | 30.23 | 31.81 | 31.80 | 32.28 | 32.26 | **32.33** | 32.27 | 32.32 | *32.42* |
|  | ×3 | 27.54 | 28.68 | 28.66 | **29.13** | 29.04 | 29.10 | 29.12 | **29.13** | *29.24* |
|  | ×4 | 26.00 | 26.88 | 26.85 | **27.33** | 27.22 | 27.28 | 27.29 | 27.30 | *27.37* |
| **B100** | ×2 | 29.32 | 30.40 | 30.44 | 30.78 | 31.13 | 31.21 | 31.16 | **31.23** | *31.31* |
|  | ×3 | 27.15 | 27.87 | 27.89 | 28.18 | 28.21 | 28.26 | 28.26 | **28.30** | *28.37* |
|  | ×4 | 25.92 | 26.51 | 26.51 | 26.77 | 26.74 | 26.77 | 26.78 | **26.81** | *26.87* |

Table-4: Results of the proposed method, compared with state-of-the-art methods on 3 datasets, in terms of PSNR (dB), with three different magnification factors (×2, ×3, ×4).

Experiment results are tabulated in Tables-4 and Tables-5, where JMPF is our proposed JMPF-based image super-resolution method, and JMPF⁻ is a trimmed version, such that the thresholds for the split-nodes are not the inherent zero-center hyperplanes, but set by the standard random-forest bagging strategy. We use the same training images (91 images) for all the algorithms as previous works [2, 4, 5, 8] do. However, for JMPF⁺, 100 more images from the General-100 dataset [22] are used, so as to check whether or not more training samples can further improve our proposed algorithm.

| **Set5(×2)** | bicubic | Zeyde[4] | ANR[2] | A+[5] | RF[8] | ARF[8] | JMPF⁻ | JMPF | **JMPF⁺** |
|---|---|---|---|---|---|---|---|---|---|
| baby | 37.05 | 38.22 | 38.42 | **38.52** | 38.47 | 38.48 | 38.40 | 38.45 | *38.45* |
| bird | 36.82 | 39.91 | 40.03 | 41.06 | 40.98 | **41.15** | 40.82 | 40.99 | *41.11* |
| butterfly | 27.43 | 30.64 | 30.54 | 32.02 | 32.27 | **32.66** | 32.58 | 32.50 | *32.79* |
| head | 34.85 | 35.62 | 35.72 | 35.82 | 35.69 | 35.73 | 35.68 | **35.73** | *35.78* |
| woman | 32.14 | 34.53 | 34.53 | 35.31 | 35.19 | 35.24 | 35.15 | **35.28** | *35.38* |
| *average* | 33.66 | 35.78 | 35.85 | 36.55 | 36.52 | **36.65** | 36.53 | 36.59 | *36.70* |

| **Set5(×3)** | bicubic | Zeyde[4] | ANR[2] | A+[5] | RF[8] | ARF[8] | JMPF⁻ | JMPF | **JMPF⁺** |
|---|---|---|---|---|---|---|---|---|---|
| baby | 33.91 | 35.13 | 35.13 | 35.23 | **35.25** | 35.15 | 35.11 | 35.16 | *35.14* |
| bird | 32.58 | 34.62 | 34.63 | **35.53** | 35.23 | 35.31 | 35.25 | 35.46 | *35.49* |
| butterfly | 24.04 | 25.93 | 25.92 | 27.13 | 27.00 | 27.39 | 27.46 | **27.48** | *27.73* |
| head | 32.88 | 33.61 | 33.64 | 33.82 | 33.73 | 33.73 | 33.72 | **33.79** | *33.76* |
| woman | 28.56 | 30.32 | 30.31 | 31.24 | 30.98 | **31.08** | 31.03 | 31.06 | *31.24* |
| *average* | 30.39 | 31.92 | 31.93 | **32.59** | 32.44 | 32.53 | 32.51 | **32.59** | *32.67* |

| **Set5(×4)** | bicubic | Zeyde[4] | ANR[2] | A+[5] | RF[8] | ARF[8] | JMPF⁻ | JMPF | **JMPF⁺** |
|---|---|---|---|---|---|---|---|---|---|
| baby | 31.78 | 33.13 | 33.07 | **33.3** | 33.26 | 33.16 | 33.09 | 33.12 | *33.12* |
| bird | 30.18 | 31.75 | 31.82 | **32.5** | 32.21 | 32.26 | 32.27 | 32.33 | *32.47* |
| butterfly | 22.10 | 23.67 | 23.58 | 24.4 | 24.32 | **24.56** | 24.55 | 24.44 | *24.63* |
| head | 31.59 | 32.23 | 32.34 | **32.5** | 32.35 | 32.37 | 32.35 | 32.45 | *32.47* |
| woman | 26.46 | 27.94 | 27.88 | **28.6** | 28.38 | 28.48 | 28.44 | 28.50 | *28.53* |
| *average* | 28.42 | 29.74 | 29.74 | **30.28** | 30.10 | 30.17 | 30.14 | 30.17 | *30.24* |

Table-5: Detailed results of the proposed method, compared with state-of-the-art methods on the dataset Set5, in terms of PSNR (dB) using three different magnification factors (×2, ×3, ×4).

Table-4 tabulates the performances, in terms of the average peak signal to noise ratio (PSNR) scores, of our proposed algorithm and other image SR methods, on the 3 datasets with different magnification factors. For the Set5 and Set14 datasets, with different magnification factors, our proposed JMPF-based algorithm can achieve a comparable performance to other recent state-of-the-art methods, such as A+

and ARF. As those random-forest-based algorithms may not be stable on small datasets, when evaluation works on extensive datasets, such as B100, our proposed algorithm JMPF can stably outperform A+ and ARF for all magnification factors (×2, ×3, ×4). Moreover, the objective quality metrics on PSNR also show that the JMPF algorithm can achieve a better performance when more samples are used for training, as shown from JMPF$^+$ in Table-4. Table-5 provides more details of the performances in datasets Set5.

To compare the visual quality of our proposed JMPF-based SR algorithm to other methods, Fig. 7, shows the reconstructed HR images using different methods. Some regions in the reconstructed images are also enlarged, so as to show the details in the images. In general, our proposed method can produce better quality images, particularly in areas with rich texture, which verifies the feature discrimination of the proposed JMPF scheme.

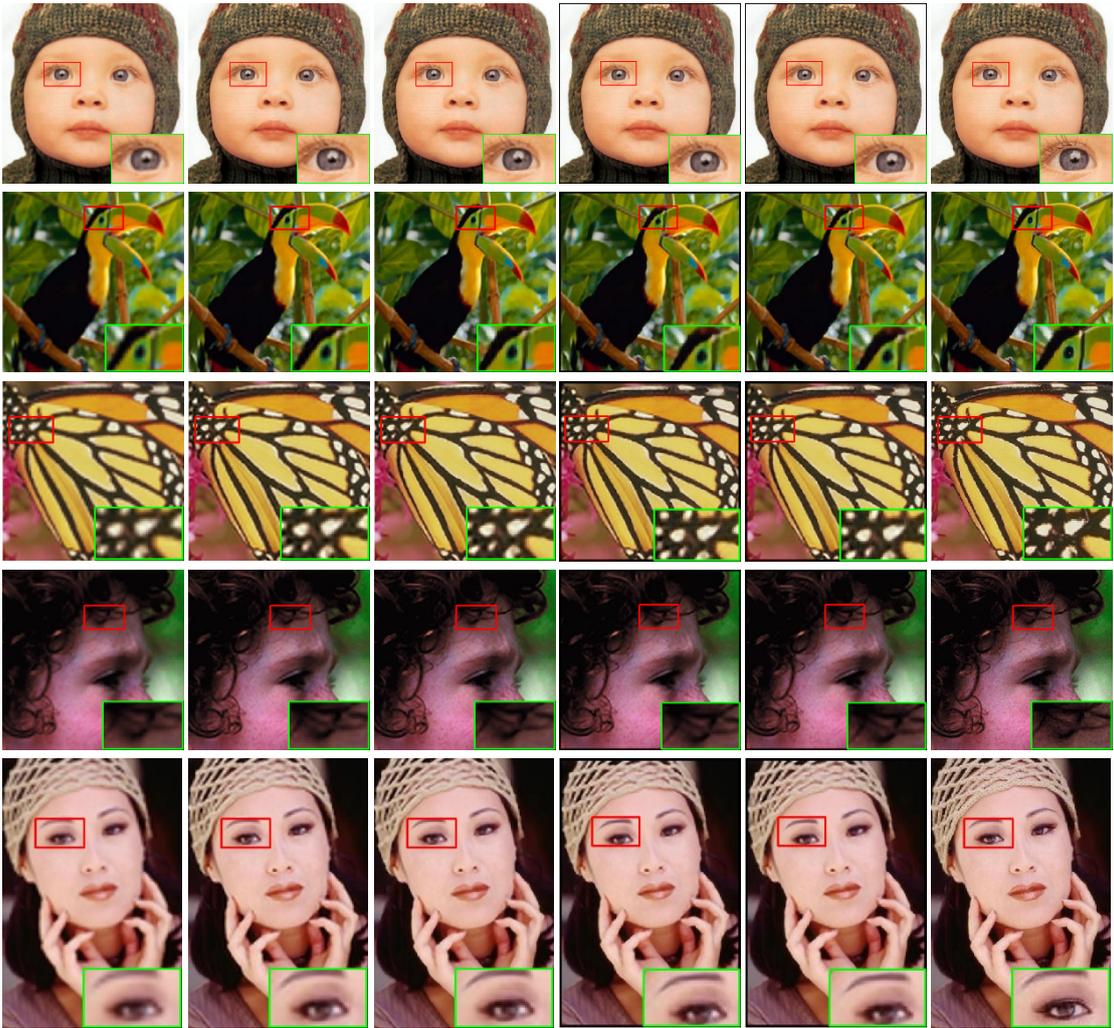

Fig. 7: Super-resolved (×3) images from Set5: (a) bicubic, (b) ANR[2], (c) A+[5], (d) ARF[8], (e) proposed algorithm JMPF, and (f) ground truth. The results show that our JMPF-based algorithm can produce more details.

## V. CONCLUSIONS

In this paper, we have proposed a novel random-forest scheme, namely the Joint Maximum Purity Forest (JMPF) scheme, which rotates the feature space into a compact, clustered feature space, by jointly maximizing the purity of all the feature-space vertices. In the new pre-clustered feature space, orthogonal hyperplanes can be effectively used in the split-nodes of a decision tree, which can improve the performance of the trained random forest. Compared to the standard random forests and the recent state-of-the-art variants, such as alternating decision forests (ADF) [23] and alternating regression forests (ARF) [24], our proposed random-forest method inherits the merits of random forests (fast training and testing, multi-class capability, etc.), and also yields promising results on both classification and regression tasks. Experiments have shown that our method achieves an average improvement of about 20% for classification and regression on publicly benchmarked datasets. Furthermore, our proposed scheme can integrate with other methods, such as ADF and ARF, to further improve the performance.

We have also applied JMPF to single-image super-resolution. We tackle image super-resolution as a clustering-regression problem, and focus on the clustering stage, which happens at the split-nodes of each decision tree. By employing the JMPF strategy, we rotate the feature space into a pre-clustered feature space, which can cluster samples into different sub-spaces more compactly in an unsupervised problem. The compact pre-clustered feature space can provide the optimal thresholds for split-nodes in decision trees, which are the zero-centered orthogonal hyperplanes. Our experiment results on intensive image benchmark datasets, such as B100, show that the proposed JMPF-based image super-resolution approach can consistently outperform recent state-of-the-art algorithms, in terms of PSNR and visual quality. Our method also inherits the advantages of random forests, which have fast speed on both the training and inference processes.